\documentclass{article}
\usepackage{ijcai25}

\usepackage{times}
\usepackage{soul}
\usepackage{url}
\usepackage[hidelinks]{hyperref}
\usepackage[utf8]{inputenc}
\usepackage[small]{caption}
\usepackage{graphicx}
\usepackage{amsmath}
\usepackage{amsthm}
\usepackage{booktabs}
\usepackage{algorithm}
\usepackage[switch]{lineno}
\usepackage{xcolor}

\usepackage{amssymb}
\usepackage{algpseudocode}
\usepackage{booktabs}  
\usepackage{multirow}  
\usepackage{makecell}  
\usepackage{adjustbox}
\usepackage{colortbl}    

\newcommand{\mn}{FreqMoE}

\title{FreqMoE: Dynamic Frequency Enhancement for Neural PDE Solvers}

\author{
Tianyu Chen$^1$
\and
Haoyi Zhou$^2$ \and
Ying Li$^3$ \and
Hao Wang$^3$ \and 
Zhenzhe Zhang$^3$ \and
Tianchen Zhu$^4$ \and \\
Shanghang Zhang$^3$ \and
Jianxin Li$^1$
\\
\affiliations
$^1$SKLCCSE, School of Computer Science and Engineering, Beihang University, China\\
$^2$School of Software, Beihang University, China\\
$^3$SKLMIP, School of Computer Science, Peking University, China\\
$^4$School of Reliability and Systems Engineering, Beihang University, China\\
\emails
\{tianyuc, haoyi, lijx\}@buaa.edu.cn
}

\begin{document}

\maketitle

\footnotetext[1]{The corresponding author is Haoyi Zhou (haoyi@buaa.edu.cn).}

\begin{abstract}
Fourier Neural Operators (FNO) have emerged as promising solutions for efficiently solving partial differential equations (PDEs) by learning infinite-dimensional function mappings through frequency domain transformations. However, the sparsity of high-frequency signals limits computational efficiency for high-dimensional inputs, and fixed-pattern truncation often causes high-frequency signal loss, reducing performance in scenarios such as high-resolution inputs or long-term predictions. To address these challenges, we propose FreqMoE, an efficient and progressive training framework that exploits the dependency of high-frequency signals on low-frequency components. The model first learns low-frequency weights and then applies a sparse upward-cycling strategy to construct a mixture of experts (MoE) in the frequency domain, effectively extending the learned weights to high-frequency regions. Experiments on both regular and irregular grid PDEs demonstrate that FreqMoE achieves up to 16.6\% accuracy improvement while using merely 2.1\% parameters (47.32$\times$ reduction) compared to dense FNO. Furthermore, the approach demonstrates remarkable stability in long-term predictions and generalizes seamlessly to various FNO variants and grid structures, establishing a new ``\textbf{L}ow frequency \textbf{P}retraining, \textbf{H}igh frequency \textbf{F}ine-tuning'' paradigm for solving PDEs.

\end{abstract}

\begin{figure}[!htbp]
    \centering
    \vspace{-0.45cm}  
    \includegraphics[width=\linewidth]{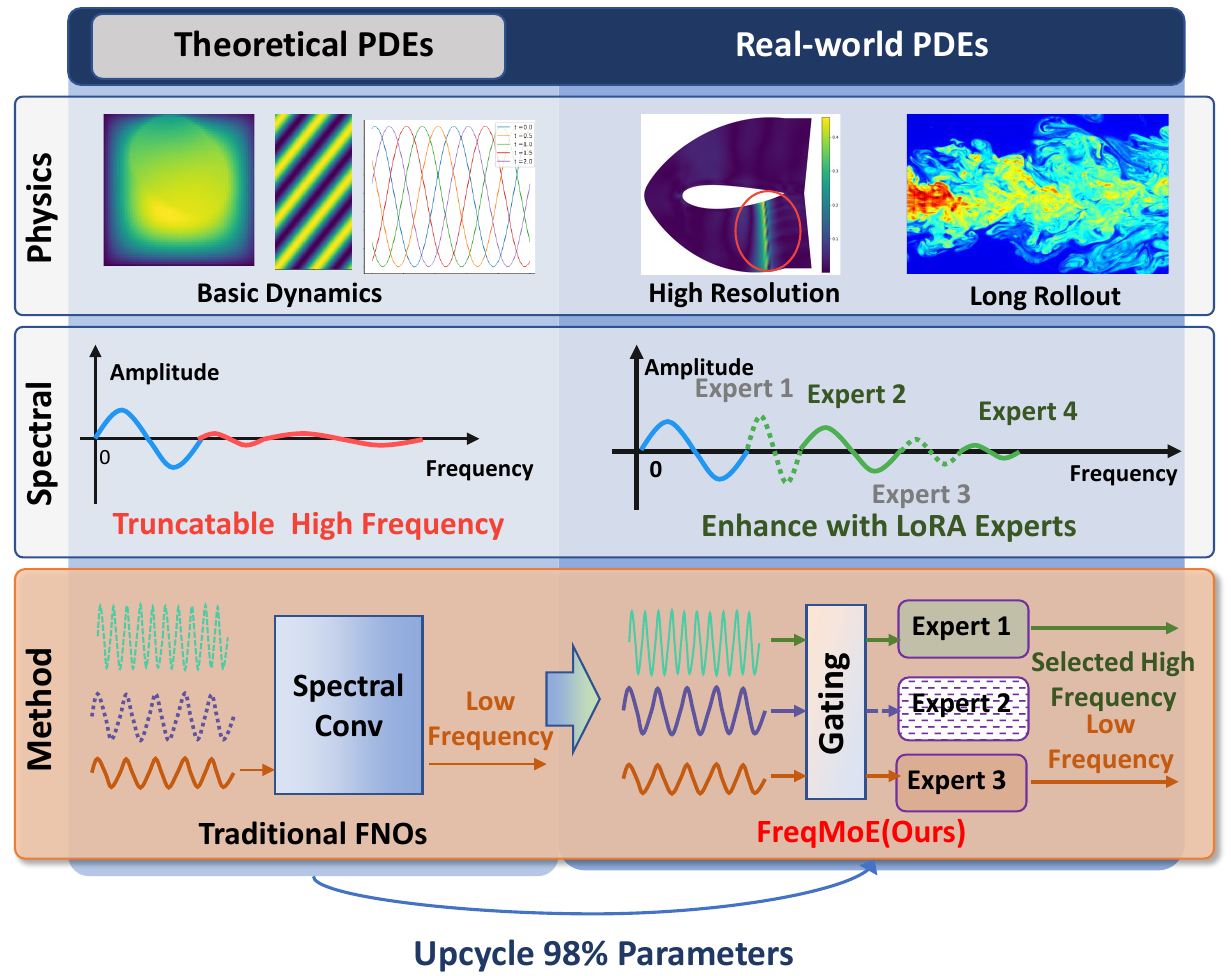}
    \caption{\textbf{Motivation of FreqMoE.} Traditional FNO directly truncates high-frequency components (left), while FreqMoE(ours) efficiently preserves them through sparse dynamic experts (right). This design enables high-frequency modeling with negligible computational overhead.}
    \vspace{-3mm}  
    \label{fig:motivation}
\end{figure}

\section{Introduction}
Efficient solutions to large-scale partial differential equations (PDEs) play a crucial role in numerous scientific computing applications, ranging from weather forecasting through Navier-Stokes equations to quantum simulations in physics~\cite{pangu-weather,fourcast,gino,quantum}. 
As spatial resolution and temporal steps increase, traditional numerical solvers face prohibitive computational costs, spurring the development of neural approaches that promise to balance accuracy with efficiency. Among these approaches, Fourier neural operators (FNO~\cite{FNO} and its variants (Geo-FNO~\cite{GeoFNO}, FFNO~\cite{FFNO}, TFNO~\cite{TFNO}) have emerged as particularly promising, leveraging the inherent sparsity of physical fields in frequency domain. By operating on a compact window of low-frequency signals while truncating higher frequencies, these methods achieve scale-free processing across arbitrary resolutions with reduced computational complexity. However, this frequency truncation presents a fundamental trade-off: while enabling computational efficiency, the loss of high-frequency information can significantly degrade performance in high-resolution scenarios and, as a result, accumulate errors in long-term predictions~\cite{PDE-refiner,Spectral-refiner}.

Recent efforts to overcome high-frequency limitations have explored post-training refinement strategies. These methods aim to recover truncated frequency information through various approaches, such as diffusion-based iterative refinement~\cite{PDE-refiner} and numerical solver guidance~\cite{Spectral-refiner}. However, existing solutions often incur substantial computational overhead or are restricted to specific scenarios, highlighting the need for a more computationally efficient approach.

To address these limitations, we propose \mn, a light-weight post-training framework inspired by upcycled MoE Models~\cite{upcycle_moe_2_into_mix_of_expert,upcycle_moe_3_bam}.
This framework enables FNO trained on low-frequency domains to adapt to high-frequency signals dynamically. Our method leverages a pre-trained FNO as a base expert for low-frequency components while initializing specialized high-frequency experts(Fig.\ref{fig:motivation}). To handle the inherent sparsity of high-frequency components, we incorporate a gating mechanism to selectively activate the most relevant high-frequency Experts during prediction. Our approach is motivated by a fundamental observation in physical systems: high-frequency signals typically exhibit strong dependencies on low-frequency components, as exemplified by the energy cascade phenomenon in fluid mechanics~\cite{energy-casacade}. Capitalizing on this physical insight, we initialize high-frequency experts re-using the base expert's weights through a LoRA-like strategy~\cite{lora}. 

Specifically, we decompose the high-frequency expert's weights into two components: a shared base weight $R_{\text{base}}$ and a low-rank delta weight $\Delta R$. 
For the $i$-th high-frequency Expert, its weights are constructed as $R_i = R +\Delta R_i$, where the shared base weights $R$ could be intialized by the pre-trained dense FNO.
This architecture offers significant computational efficiency improvement through two key design choices: (1) during post-training, low-rank delta weights are parameter-efficient, and (2) during inference, only the Top-K selected experts participate in prediction. The extremely low-rank nature of $\Delta R$ ensures minimal computational overhead in both stages, making our method particularly practical for real-world applications.


Through extensive evaluation on both regular and irregular grid PDEs, we demonstrate the effectiveness of FreqMoE in high-resolution and long-term prediction scenarios. Our experiments reveal compelling advantages: in high-resolution tasks (512×512), FreqMoE achieves up to 16.6\% accuracy improvement while using merely 2.1\% parameters (47.32$\times$ reduction) compared to conventional FNO. This efficiency extends to unstructured meshes, where FreqMoE maintains superior performance with 27.37$\times$ parameter reduction. Furthermore, long-term rollout experiments showcase FreqMoE's stability in mitigating error accumulation, particularly in challenging high-resolution scenarios.

The key contributions of this work are threefold:
\begin{enumerate}
    \item We propose FreqMoE, a lightweight post-training framework that dynamically enhances high-frequency processing capabilities in neural PDE solvers. Our approach generalizes seamlessly across the FNO family on both structured and unstructured grids, establishing an efficient "low-frequency pretraining, high-frequency fine-tuning" paradigm.
    \item Inspired by physical principles of frequency dependencies in PDEs, we develop a LoRA-based expert initialization scheme that efficiently reuses low-frequency weights. This design achieves remarkable parameter efficiency (47.32$\times$ reduction) while maintaining competitive performance through sparse dynamic computation.
    \item Through comprehensive evaluation on diverse PDE systems, we demonstrate that FreqMoE significantly outperforms conventional FNO variants, achieving up to 16.6\% accuracy improvement in high-resolution tasks (512×512) and superior stability in long-term predictions, all while maintaining minimal computational overhead.
\end{enumerate}

\section{Related Works}

\begin{figure*}[htbp]
    \centering
    \includegraphics[width=\linewidth]{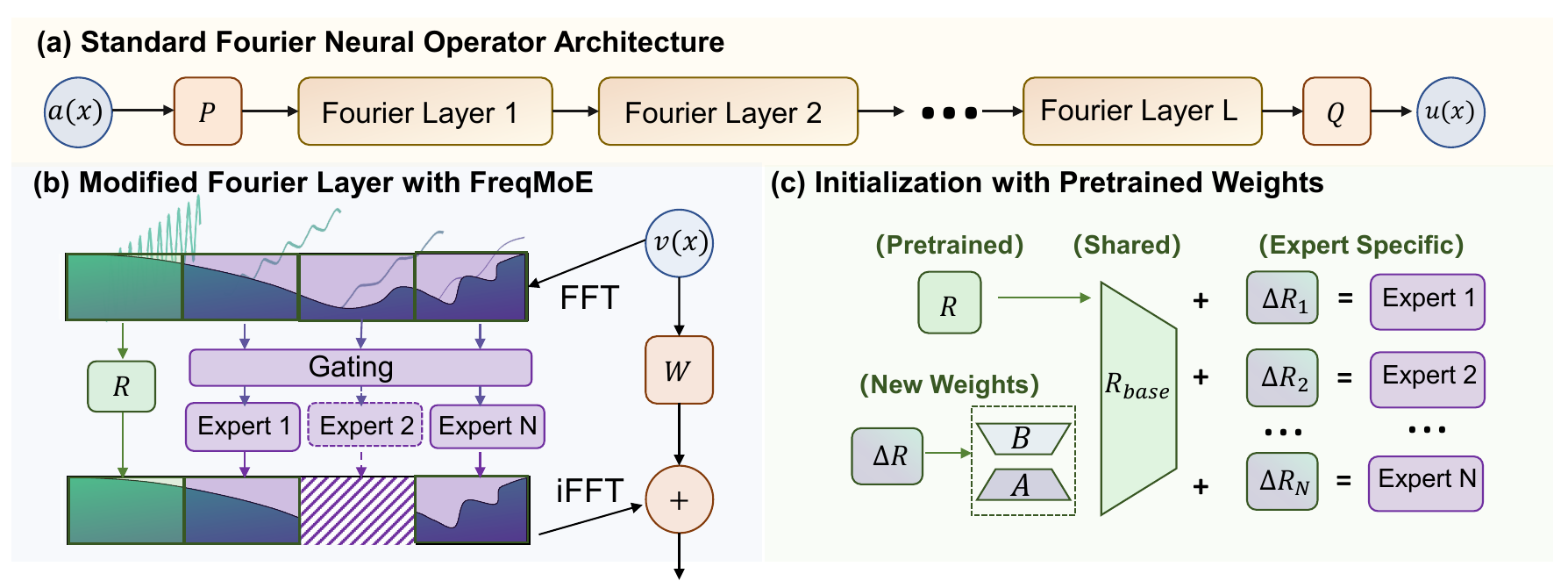}
    \caption{\textbf{Methods overview of \mn.} \textbf{(a) The standard Fourier Neural Operator (FNO) architecture}  consisting of input lifting (P), a sequence of Fourier layers, and output projection (Q). 
    \textbf{(b) Our modified Fourier layer design with a mixture-of-experts mechanism}, where the gating networker dynamically assigns frequency components to  specialized experts after FFT decomposition. High-frequency components (lighter shades) are processed by high-frequency experts, while low-frequency components are handled by the base expert. 
    \textbf{(c) Our expert initialization strategy}, where pre-trained weights $R$ are used as a shared base component $R_\text{base}$ and  expert-specific delta weights $\Delta R$ are initialized with LoRA trick, enabling efficient parameter sharing and specialized frequency processing.}
    \label{fig:methods}
  \end{figure*}

\subsection{Fourier Neural Operators}

Fourier Neural Operators (FNO)~\cite{FNO} have revolutionized PDE solving by introducing FFT-based spectral convolution layers to learn mappings between infinite-dimensional function spaces. This foundational work has sparked numerous architectural innovations: Geo-FNO~\cite{GeoFNO} and SFNO~\cite{SFNO} extended the framework to handle irregular grids and spherical geometries, while F-FNO~\cite{FFNO} enhanced scalability through separable spectral convolutions and advanced training strategies. T-FNO~\cite{TFNO} further improved parameter efficiency and generalization by implementing global tensor decomposition. 
Despite these advancements, the issue of high-frequency truncation—a critical limitation in FNO—remains largely unaddressed. 
Our work directly tackles this gap by enhancing high-frequency signal processing capabilities, offering a complementary approach that integrates seamlessly with existing FNO architectures to further improve performance.


\subsection{Sparse Upcycling Techniques}
Sparse upcycling has emerged as a powerful paradigm for efficient model enhancement, leveraging sparsely activated Mixture-of-Experts (MoE) initialized from pre-trained dense models. This approach has demonstrated remarkable success across diverse domains, from language models (T5)~\cite{upcycle_moe_t5_vit} to vision-language systems (MoE-LLAVA)~\cite{moe_llava} and medical applications (MoE-Med)~\cite{Med-MoE}, consistently outperforming sparse models trained from scratch while significantly reducing computational costs. 
Building on these principles, our work pioneers the application of sparse upcycling to frequency-domain learning. 
We introduce a novel framework that utilizes pre-trained FNO to efficiently enhance high-frequency components, achieving improved performance with minimal additional training overhead.

\section{Method}

FreqMoE extends FNO with dynamic frequency processing through a lightweight expert system. As shown in Fig. \ref{fig:methods}, our framework splits the frequency spectrum into chunks and processes high-frequency components via specialized experts derived from pre-trained FNO. We first introduce the frequency-domain MoE design (Sec. \ref{Sec:MoE}), then present our efficient expert initialization scheme (Sec. \ref{sec:upcycle}), followed by the training strategy that enables sparse computation. We begin by reviewing the basics of FNO and MoE systems.

\subsection{Preliminary}

\textbf{Neural PDE Solvers with FNO.} Fourier Neural Operator (FNO) learns a parameterized operator $\mathcal{G}_\theta$ that maps input functions to output solutions in infinite-dimensional spaces. The core of FNO is its Fourier layer(Fig.\ref{fig:methods}(a)), which performs spectral convolution through: $\mathcal{K}^{(l)}(z^{(l)}) = \text{IFFT}(R^{(l)} \cdot \text{FFT}(z))$, where $R^{(l)} \in \mathbb{C}^{H \times H \times M_1... M_d}$ are learnable weights operating on truncated frequency modes $\{M_{(i)}| i \in \{1,2,...d\}\}$. This frequency truncation, while computationally efficient, leads to information loss in high-frequency components.

\textbf{Mixture-of-Experts (MoE).} A standard MoE layer consists of a gating network $\mathcal{P}_\theta$ and $N$ expert networks $E_{\theta_j}$, computing outputs as: $\text{MoE}(x) = \sum \limits_{j=1}^{N}\text{TopK}(\text{Softmax}(\mathcal{P}_\theta(x)_j)) \cdot E^{(j)}_\theta(x)$. In our frequency-domain adaptation, experts specialize in different frequency chunks, with the gating network determining the activation of high-frequency computations during inference.

\subsection{MoE in Frequency Domain}
\label{Sec:MoE}

Traditional FNO truncates high frequencies for efficiency, but this fixed cutoff limits model capacity. Our FreqMoE design addresses this limitation by adaptively processing the frequency spectrum based on two observations: high-frequency signals in PDEs are naturally sparse, and their patterns are often localized. These properties make the frequency domain particularly suitable for expert-based processing.

\textbf{Frequency Domain Partitioning.} In the standard spectral convolution, for an input feature map $z \in \mathbb{R}^{S}$, its Fourier transform $\hat{z} = \text{FFT}(z) \in \mathbb{C}^{S}$ is truncated to retain only the lowest frequency bands for processing: $o_P = R_\theta \cdot \hat{z}_P$, where $R_\theta$ represents the learnable weights. We generalize this fixed truncation scheme by partitioning the frequency spectrum into $J = S/P$ bands:
$\{\hat{z}_P^{(i)} \in \mathbb{C}^P \mid i = 0,1,\ldots,J-1\}$
where bands are ordered by increasing frequency, with $\hat{z}_P^{(0)}$ containing the lowest frequency components.

\textbf{Expert Specialization.} We assign $N$ specialized experts $\{E_{\theta_i} \mid i = 1,\ldots,N\}$ ($N \leq J-1$) to process different high-frequency bands, while keeping the original FNO weights $R_\theta$ as the base expert for low-frequency components $\hat{z}_P^{(0)}$. This design stems from a key insight in PDE solutions: low frequencies capture global patterns that require careful processing, while high frequencies reflect local details that can benefit from specialized, targeted handling.

\textbf{Adaptive Frequency Gating.} To exploit the natural sparsity in high-frequency signals, we design a gating mechanism $g_\theta$ that selectively activates experts based on frequency content:
$g_\theta(\hat{z}_P^{(i)}) = \sigma(\frac{w_\theta \cdot \hat{z}_P^{(i)}}{\tau})$
where $\tau$ is a temperature parameter and $\sigma$ is the sigmoid function. The forward computation follows:
\begin{equation}
o_P^{(i)} = \begin{cases}
R_\theta(\hat{z}_P^{(i)}), & i = 0 \\
g_\theta(\hat{z}_P^{(i)}) \cdot E_{\theta_{i}}(\hat{z}_P^{(i)}), & i = 1,\ldots,N \\
0, & \text{otherwise}
\end{cases}
\end{equation}

To encourage sparse expert utilization during training, we add a sparsity loss on gate values:
\begin{equation}
\mathcal{L}_{\text{sparse}} = \mathbb{E}[\sum_{i=1}^N g_\theta(\hat{z}_P^{(i)})].
\end{equation}

This regularization pushes the model to activate only the most relevant experts for each frequency band, leading to more efficient inference.

\textbf{Inference-Time Sparsity.} During inference, we leverage the sparse nature of expert utilization by activating only the top-K experts ($K \leq N$) based on their gating values:
$\text{active\_experts} = \text{TopK}({g_\theta(\hat{z}_P^{(i)})}{i=1}^N, K)$.

The inference computation becomes:
\begin{equation}
o_P^{(i)} = \begin{cases}
R_\theta(\hat{z}_P^{(i)}), & i = 0 \\
g_\theta(\hat{z}_P^{(i)}) \cdot E_{\theta_{i}}(\hat{z}_P^{(i)}), & i \in \text{active\_experts} \\
0, & \text{otherwise.}
\end{cases}
\end{equation}

This sparse activation strategy significantly reduces computational overhead while maintaining model performance, as high-frequency components typically require selective rather than comprehensive processing.
The final output in the spatial domain is obtained through the inverse Fourier transform: $o = \text{IFFT}({o_P^{(i)}}_{i=0}^{J-1})$, where the unprocessed high-frequency components are naturally zero-padded.

\subsection{Sparsely Upcycle the Low-frequency Weight}
\label{sec:upcycle}

After establishing the expert structure, a key challenge is how to efficiently initialize these experts. Instead of training from scratch, we propose a sparse upcycling strategy (Algorithm \ref{alg:sparse_upcycle}) that leverages pre-trained FNO weights while keeping the parameter count low. This approach allows experts to inherit low-frequency patterns while developing specialized high-frequency processing capabilities.

\textbf{Parameter-Efficient Weight Adaptation.} For each expert $E_{\theta_i}$, we decompose its adapted weights $R_{\theta_i}$ into a shared base component and an expert-specific delta:

\begin{equation}
R_{\theta_i} = R_{\theta} + \Delta R_{\theta_i},
\end{equation}

where $R_{\theta} \in \mathbb{C}^{H \times H \times M_1 \times ... \times M_d}$ represents the pre-trained weights in the low-frequency domain. The expert-specific adaptation $\Delta R_{\theta_i}$ is computed through low-rank decomposition:

\begin{equation}
\Delta R_{\theta_i} = \alpha \cdot A_{\theta^i}B_{\theta^i}
\end{equation}

with $A^{(i)} \in \mathbb{C}^{r \times H}$ and $B_{\theta^i} \in \mathbb{C}^{H \times r \times M_1 \times ... \times M_d}$ being low-rank adaptation matrices with rank $r \ll H$. This formulation reduces the adaptation parameters from $O(H^2\prod \limits_{i}^{d} M_{(i)})$ to $O(rH(1 + \prod \limits_{i}^{d} M_{(i)}))$ per expert.

\begin{algorithm}
    \caption{Sparsely Upcycling of FNO}
    \label{alg:sparse_upcycle}
    \begin{algorithmic}[1]
        \Require Pretrained FNO Model F, number of experts $N$, rank $r$, scaling factor $\alpha$
        \Ensure Upcycled FreqMoE Model
        \State // Initialize expert parameters
        \For{each Fourier layer $l$}
            \State $R^{(l)}_\theta \leftarrow \text{F.get\_pretrained\_weights}(l)$
            \State Initialize gating network $g^{(l)}_\theta $
            \For{i in 1 to N}
                \State Initialize expert $R^{(l)}_{\theta^i} \leftarrow R^{(l)}_\theta + \alpha \cdot A^{(l)}_{\theta^i} B^{(l)}_{\theta^i}$
            \EndFor
        \EndFor
        \State \Return{New FreqMoE Model F}
    \end{algorithmic}
\end{algorithm}


\subsection{Bridging Low and High Frequency Learning}

FNO effectively addresses the challenge of learning in infinite function spaces, yet it primarily captures low-frequency patterns as resolution increases. This aligns well with PDE characteristics where dominant features reside in low frequencies, making it efficient to learn fundamental patterns. FreqMoE builds upon this insight by establishing a bridge between low and high frequencies through expert upcycling, enabling pattern transfer across frequency bands.

This design naturally leads to an efficient learning paradigm: \textit{Low-Frequency Pretraining, High-Frequency Fine-tuning(LPHF}). Since inference over high-resolution PDE solutions is computationally expensive, LPHF allows us to learn core patterns from abundant low-resolution data, then adapt to high frequencies with much fewer parameters. As demonstrated in our experiments on both regular (CFD) and irregular (AirFoil) grids, this paradigm significantly accelerates neural operator inference while maintaining high accuracy across resolutions.

\begin{table*}[htbp]
\caption{\textbf{Performance on Regular-Grid PDEs.} Comparison of models with varying frequency modes, where \# Params indicates the number of parameters activated during inference. \underline{Underlined} values represent the best performance achieved by FNO baselines. Results with blue background show our FreqMoE, where superscript $^*$ and $^\dagger$ denote models trained from scratch and upcycled from dense FNO, respectively. The \textbf{bold} values highlight our best performance.}
\renewcommand{\arraystretch}{1.0}
\centering
\label{tab:fno_results}
\begin{tabular}{c|cc|ccc}
\toprule
                                                                                                              &                                  &                                                   & \multicolumn{3}{c}{\textbf{Relative L2 Error(L2RE)$\downarrow$}}                  \\
\multirow{-2}{*}{\textbf{Models}}                                                                             & \multirow{-2}{*}{\textbf{Modes}} & \multirow{-2}{*}{\textbf{\# Params $\downarrow$}} & \textbf{CFD-Rand 128} & \textbf{CFD-Rand 512} & \textbf{CFD-Turb 512} \\ \midrule
                                                                                                              & (4,4)                              & 142.69K                                           & 0.0481 $\pm$ 0.0061& 0.3856 $\pm$ 0.0434& 0.2445 $\pm$ 0.0259\\
                                                                                                              & (16,16)                            & 2.11M                                             & 0.0434 $\pm$ 0.0052& 0.3981 $\pm$ 0.0434& \underline{0.2164} $\pm$ 0.0316\\
\multirow{-3}{*}{\textbf{\begin{tabular}[c]{@{}c@{}}FNO\\ (Dense)\end{tabular}}}                              & (32,32)                           & 8.40M                                             & \underline{0.0410}$\pm$ 0.0045& \underline{0.3742} $\pm$ 0.0427& 0.2436 $\pm$ 0.0267\\ \midrule
\rowcolor{cyan!10} 
\cellcolor{cyan!10}                                                                                      & (32,32)$^*$        & 177.53K                                           & 0.0404$\pm$ 0.0047& 0.3720 $\pm$ 0.0469& 0.2320 $\pm$ 0.0264\\
\rowcolor{cyan!10} 
\multirow{-2}{*}{\cellcolor{cyan!10}\textbf{\begin{tabular}[c]{@{}c@{}}FreqMoE\\ (Sparse)\end{tabular}}} & (4,4)$\rightarrow$(32,32)$^\dagger$   & 177.53K                                           & \textbf{0.0370} $\pm$ \textbf{0.0038}& \textbf{0.3122} $\pm$ \textbf{0.0257}& \textbf{0.1934} $\pm$ \textbf{0.0226}\\ \midrule
\rowcolor{gray!10}
\multicolumn{3}{c|}{\textbf{Params Reduction}} & \multicolumn{3}{c}{\textcolor{blue}{\textbf{$\downarrow$ 47.32$\times$}}} \\ 
\bottomrule
\end{tabular}
\end{table*}

\section{Experiments}

We conduct systematic evaluations of FreqMoE across two critical scenarios demanding effective high-frequency modeling: high-resolution inputs and long-term prediction rollouts. Our experimental framework follows a progressive approach: (1) Training base FNO models on low-frequency regimes, then (2) transforming them into sparse FreqMoE architectures through our parameter-efficient upcycling strategy (Section \ref{sec:upcycle}). We measure model effectiveness through activated parameter counts(\# Params) and prediction accuracy (L2 relative error), with comprehensive ablation studies on frequency adaptation mechanisms.

\subsection{Datasets}
To demonstrate FreqMoE's versatility across different PDE domains and discretization schemes, we select benchmark problems from both regular and irregular grid settings. We include the detail description of PDE problems in Appendix~\ref{app:problem_description}.

\textbf{Regular-grid PDEs.} From PDEbench~\cite{PDEbench}, we choose vortex-dominated flows under Random and Turbulent initializations. Evaluations at 128×128(CFD-Rand 128) and 512×512(CFD-Rand 512, CFD-Turb 512) resolutions test progressive frequency handling capabilities, where higher resolutions reveal finer turbulent structures.


\textbf{Irregular-grid PDEs.} Using Geo-FNO's~\cite{GeoFNO} challenging scenarios: (1)\textit{Airfoil}, transonic flows over parameterized NACA-0012 airfoils (Mach 0.8) with shock-induced high frequencies on adapted C-grids (~200×50). 
(2)\textit{Elasticity}, nonlinear material deformations with central voids (radius 0.2-0.4), modeled via ~1000 FEM nodes capturing stress concentrations.

\subsection{Baseline and Implementation}

We evaluate FreqMoE against two strong FNO variants, with implementation details summarized below and expanded in Appendix~\ref{app:implentation_details}.

\textbf{FNO Baselines.} For regular grids, we implement vanilla FNO~\cite{FNO} with four Fourier layers (width=32) under three spectral configurations: (4,4), (16,16), and (32,32) modes. Inputs include velocity components (Vx, Vy), pressure, and density fields. Training adopts single-step prediction with Adam optimizer (initial lr=0.001).

\begin{figure}[!hbp]
    \centering
    \vspace{-0.5cm}
    \includegraphics[width=0.81\linewidth]{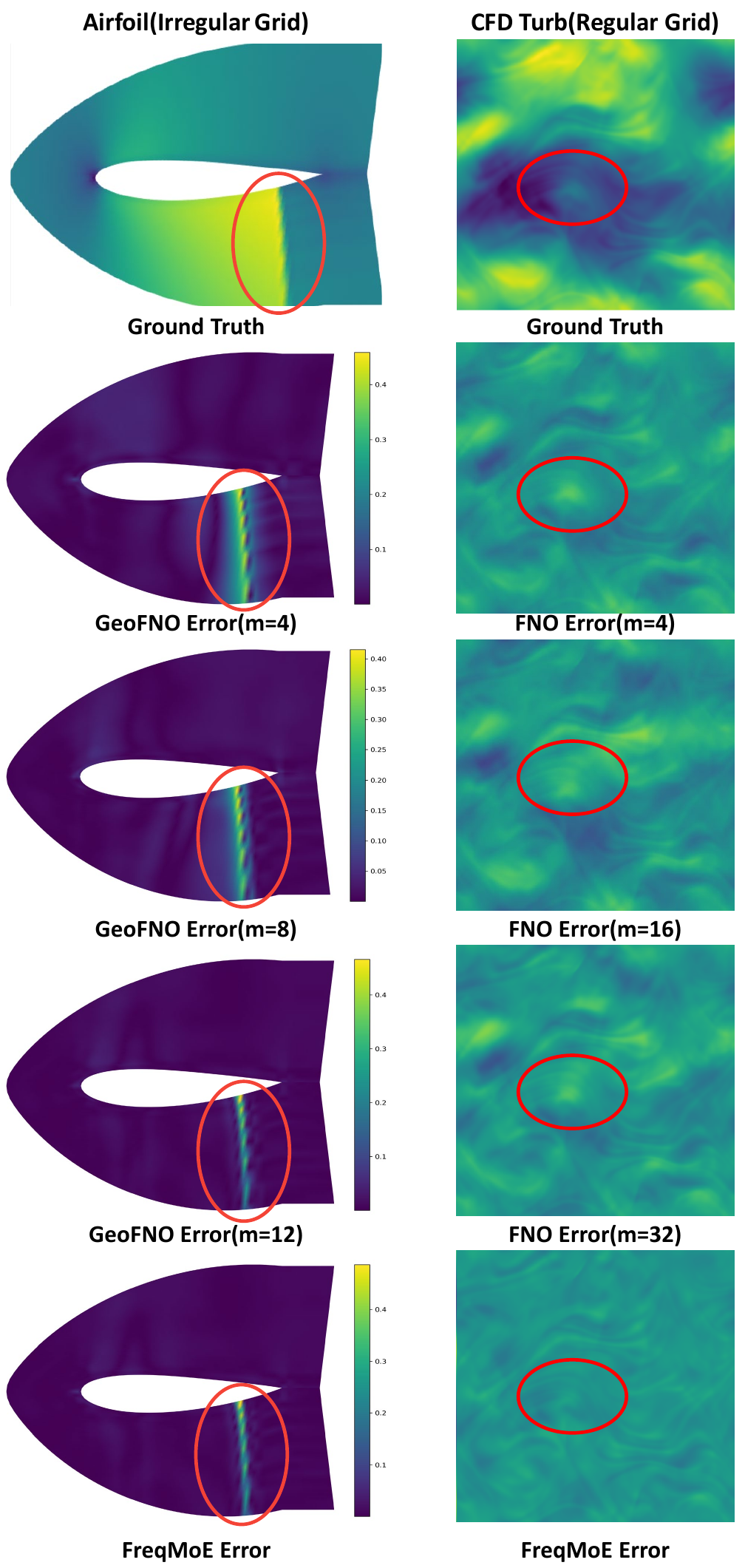}
    \caption{\textbf{Visualization of prediction errors.} \textbf{Left Column}: Irregular Grid Results from AirFoil. \textbf{Right Column}: Regular Grid Results from CFD-Turb 512. Red circles highlight regions with high-frequency components, where our FreqMoE demonstrates better capability in capturing fine-grained spatial details compared to  FNO.}
    \label{fig:case_study}
\end{figure}

\textbf{GeoFNO Baselines. } For irregular grids, we extend Geo-FNO~\cite{GeoFNO} with task-specific designs:(1)\textit{Airfoil}, asymmetric modes (2,4) to (16,32) capture shock waves, using 4 input channels (coordinates + physical fields). (2)\textit{Elasticity}, symmetric modes (2,2) to (16,16) model stress concentrations, enhanced with polar coordinate encoding. Both variants employ the IPHI module for coordinate transformation, trained with 50\% learning rate decay every 50 epochs.

\begin{table*}
\renewcommand{\arraystretch}{1.1}
\centering
\caption{\textbf{Performance on Irregular-Grid PDEs.} Comparison of models on two representative irregular-grid tasks: AirFoil and Elasticity, where \# Params indicates the number of parameters activated during inference. \underline{Underlined} values represent the best performance achieved by Geo-FNO baselines. Results with blue background show our FreqMoE approach, where superscript $^*$ and $^\dagger$ denote models trained from scratch and upcycled from dense Geo-FNO, respectively. The \textbf{bold} values highlight our best performance.}
\label{tab:geofno_results}
\begin{tabular}{c|ccc|ccc}
\toprule
                                                                                                              & \multicolumn{3}{c|}{\textbf{AirFoil}}                                                              & \multicolumn{3}{c}{\textbf{Elasticity}}                                                       \\ 
\multirow{-2}{*}{\textbf{Models}}                                                                             & \textbf{Modes}                 & \textbf{\# Params$\downarrow$} & \textbf{L2RE$\downarrow$} & \textbf{Modes}            & \textbf{\# Params$\downarrow$} & \textbf{L2RE$\downarrow$} \\ \midrule
                                                                                                              & (2,4)                            & 74.27k                                & 0.0270 $\pm$ 0.0038& (2,2)                       & 49.06K                                & 0.0236 $\pm$ 0.0034\\
                                                                                                              & (4,8)                            & 270.88k                               & 0.0161 $\pm$ 0.0020& (4,4)                       & 171.94k                               & 0.0386 $\pm$ 0.0057\\
                                                                                                              & (8,16)                           & 1.06M                                 & 0.0153 $\pm$ 0.0016& (8,8)                       & 663.46k                               & 0.0312 $\pm$ 0.0037\\
                                                                                                              & (12,24)                         & 2.37M                                 & \underline{0.0152} $\pm$ 0.0016& (12,12)                     & 1.48M                                 & \underline{0.0229} $\pm$ 0.0025\\
\multirow{-5}{*}{\textbf{\begin{tabular}[c]{@{}c@{}}Geo-FNO\\ (Dense)\end{tabular}}}                          & (16,32)                          & 4.20M                                 & 0.0708 $\pm$ 0.0100& (16,16)                     & 2.62M                                 & 0.0540 $\pm$ 0.0067\\ \midrule
\rowcolor{cyan!10}
\cellcolor{cyan!10}                                                                                      & (16,32)$^*$                 & 148.06k                               & 0.0432 $\pm$ 0.0038& (16,16)$^*$            & 94.57k                                & 0.0397 $\pm$ 0.0046\\
\rowcolor{cyan!10} 
\multirow{-2}{*}{\cellcolor{cyan!10}\textbf{\begin{tabular}[c]{@{}c@{}}FreqMoE\\ (Sparse)\end{tabular}}} & (2,4)$\rightarrow$(16,32)$^\dagger$  & 148.06k                               & \textbf{0.0154} $\pm$ 0.0013& (2,2)$\rightarrow$(16,16)$^\dagger$  & 94.57k                                & \textbf{0.0217} $\pm$ 0.0018\\ \midrule
\rowcolor{gray!10}
\multicolumn{1}{c|}{\textbf{Params Reduction}} & \multicolumn{3}{c|}{\textcolor{blue}{\textbf{$\downarrow$ 27.37$\times$}}} & \multicolumn{3}{c}{\textcolor{blue}{\textbf{$\downarrow$ 26.70$\times$}}} \\
\bottomrule
\end{tabular}
\end{table*}

\textbf{FreqMoE Configuration.}  Our architecture introduces two key innovations: (1)\textit{Dynamic Expert Selection}, expands spectral capacity from (4,4)→(32,32) for regular grids and (2,4)→(16,32)/(2,2)→(16,16) for irregular grids, activating only Top-2 experts during inference. (2) \textit{Upcycling Strategy}, initializes weights from pre-trained base models via low-rank factorization (rank=4), contrasted with scratch training. Training stabilizes via expert sparsity loss (factor $\alpha=0.1$) with identical hyperparameters to baselines for fair comparison. This design achieves parameter efficiency while preserving high-frequency resolution – critical for our later analyses of activation patterns and long-term stability.

\subsection{Comprehensive Evaluation and Insights}
Our experiments systematically validate FreqMoE's capabilities through two analytical lenses: (1) The post-training performance improvement in relative L2 error. (2) The inference efficiency improvement via activated parameters reduction. Key findings reveal that FreqMoE achieves superior high-frequency modeling with 6-28× parameter reduction compared to dense counterparts, while maintaining robust performance in all scenarios.


\textbf{High-Resolution Regular Grid Analysis.} Our experiments reveal fundamental limitations in conventional FNO's frequency scaling approach. As shown in Table~\ref{tab:fno_results}, naively expanding FNO from (4,4) to (32,32) modes yields diminishing returns - while the 32×32 model achieves marginal gains on 128×128 resolution (4.81\%$\rightarrow$4.10\% L2RE), it degrades performance on high-resolution CFD-Turb 512 (24.45\%$\rightarrow$24.36\%) despite 59$\times$ parameter growth. This exposes a critical tradeoff: dense spectral models over-parameterize high-frequency components that rarely activate in practice.

FreqMoE breaks this tradeoff through dynamic expert specialization. With only 177.53K active parameters (47$\times$ fewer than (32,32) FNO), our sparse model reduces errors by 9.8\%-16.6\% across resolutions. The upcycled variant (4$\rightarrow$32 modes) achieves particularly striking improvements: 20.6\% error reduction on CFD-Turb 512 compared to its dense counterpart, demonstrating superior turbulence modeling. Spatial error maps in Figure~\ref{fig:case_study} validate this behavior - while dense FNO accumulates errors in vortex cores (red circles), FreqMoE maintains accurate predictions through adaptive frequency allocation.This resolution-aware adaptation explains FreqMoE's dual advantage: preserving low-frequency stability (4.10\%$\rightarrow$3.70\% on CFD-Rand 128) while capturing high-frequency details (24.36\%$\rightarrow$19.34\% on CFD-Turb 512). 

\begin{figure}
    \centering
    \includegraphics[width=\linewidth]{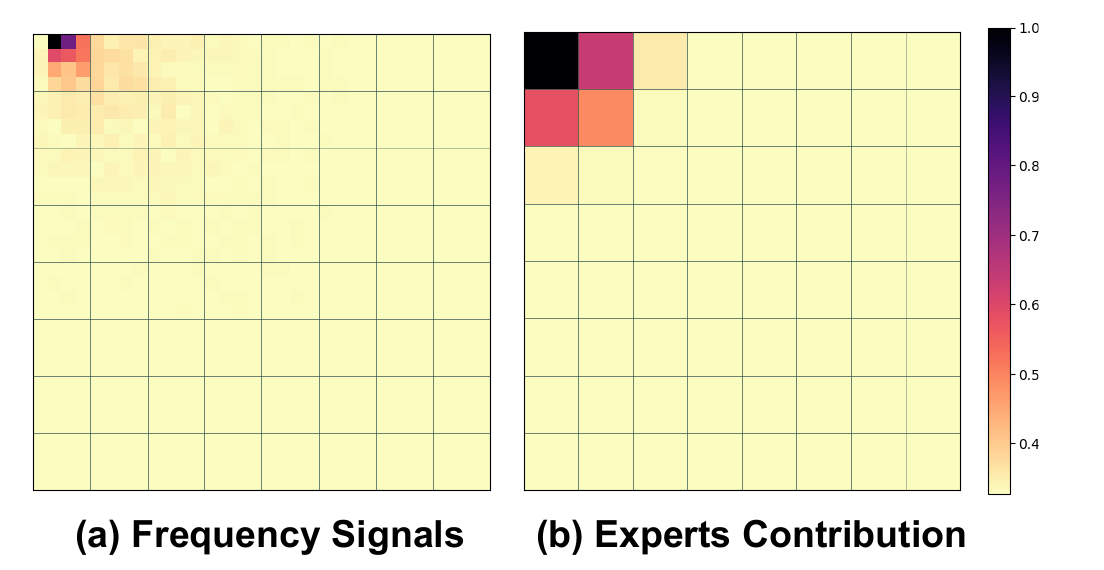}
    \caption{\textbf{Visualization of Experts.} (a) Distribution of frequency signals after FFT transformation. (b) Activation patterns of experts in FreqMoE, where each grid cell represents a frequency mode chunk. Beyond capturing low-frequency signals in the top-left corner, FreqMoE dynamically activates experts to capture surrounding high-frequency components.}
    \label{fig:viz_experts}
\end{figure}

\begin{figure*}[htbp]
    \centering
    \includegraphics[width=1.0\linewidth]{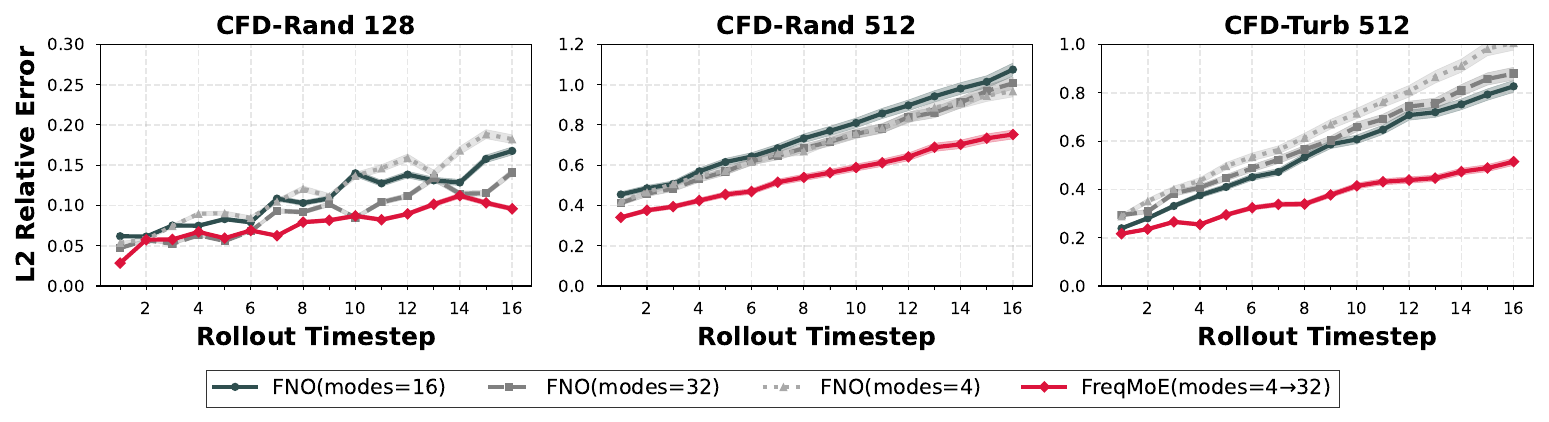}
    \caption{\textbf{Long-term Prediction Performance on Different CFD Datasets.} The plots show the L2 relative error evolution during rollout prediction across three datasets of varying complexity. FreqMoE demonstrates superior stability in long-term predictions compared to baseline FNO models with different mode configurations. This advantage becomes particularly pronounced in high-resolution scenarios (CFD-Rand 512 and CFD-Turb 512), where the error growth is significantly moderated.}
    \label{fig:rollout}
\end{figure*}

\textbf{Results on Irregular-Grid PDEs.} The challenges of irregular grids exacerbate conventional Geo-FNO methods' inefficiency in high-frequency processing. As Table~\ref{tab:geofno_results} demonstrates, naively expanding Geo-FNO to (16,32) modes for AirFoil catastrophically degrades performance (L2RE surges from 0.0152 to 0.0708) despite 4.2M parameters - revealing dense models' vulnerability to spectral over-parameterization. FreqMoE addresses this through sparse high-frequency specialization: with merely 148K parameters (28$\times$ fewer than (16,32) modes Geo-FNO), our model achieves near-identical AirFoil accuracy (0.0154 vs 0.0152) while reducing Elasticity errors by 5.2\%. This efficiency stems from dynamic frequency enhancement - preserving critical high-frequency components around geometric discontinuities (airfoil edges in Figure~\ref{fig:case_study}) without parameter bloat. The 26.7$\times$ parameter reduction in Elasticity tasks particularly highlights FreqMoE's advantage in handling stress concentration areas where high-frequency signals dominate.

\textbf{Sparsely Activation of Experts.} As described in Section \ref{Sec:MoE}, FreqMoE dynamically activates high-frequency experts based on input signals through its gating mechanism. Figure \ref{fig:viz_experts} illustrates both the frequency distribution and expert activation patterns. The frequency visualization (Figure \ref{fig:viz_experts}(a)) reveals that high-frequency components in PDE solutions exhibit natural sparsity, with signal energy primarily concentrated in the low-frequency region (top-left corner). The expert activation map (Figure \ref{fig:viz_experts}(b)) demonstrates how FreqMoE's gating mechanism responds to this spectral characteristic - while maintaining consistent engagement with low-frequency experts, it selectively activates high-frequency experts only when corresponding signal components are present. This adaptive activation pattern suggests that FreqMoE can effectively identify the sparse high-frequency patterns while preserving computational efficiency through targeted expert utilization.

\textbf{Analysis of Rollout Performance.} The rollout experiments demonstrate ~\mn's effectiveness in mitigating error accumulation during long-term predictions. In low-resolution scenarios (CFD-Rand 128), all models show relatively stable performance, with FreqMoE maintaining a slight edge in accuracy. However, the advantages of FreqMoE become substantially more evident in high-resolution cases. For CFD-Rand 512, while baseline FNO models exhibit rapid error accumulation regardless of their mode numbers, FreqMoE maintains a significantly lower error trajectory, with 31.67\% reduction in final prediction error. This pattern is further amplified in the more challenging CFD-Turb 512 dataset, where turbulent flows introduce additional high-frequency components. Here, ~\mn's adaptive frequency modeling capability proves particularly valuable, effectively containing error growth even as prediction steps extend. This performance gap suggests that FreqMoE's dynamic expert activation successfully preserves critical high-frequency information that traditional FNO models typically lose, thereby preventing the cascade of prediction errors in complex fluid simulations.

\begin{figure}[htbp]
    \centering
    \includegraphics[width=\linewidth]{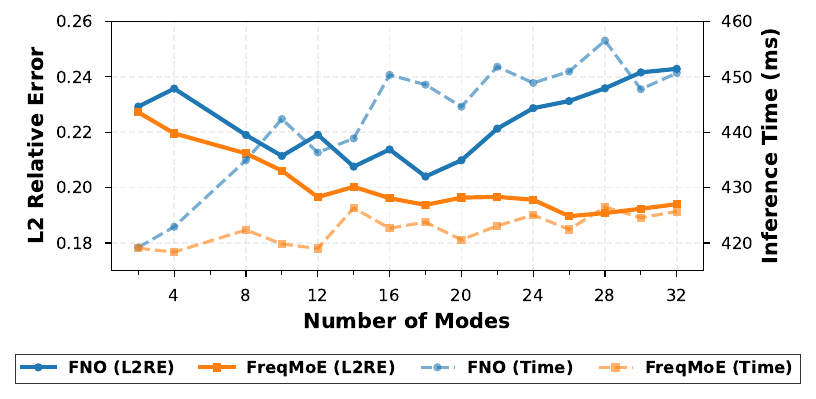}
    \caption{\textbf{Performance and Efficiency across different frequency modes.} The solid lines (left y-axis) show the L2 Relative Error (L2RE) achieved by different numbers of modes, while the dashed lines (right y-axis) represent the corresponding inference time measured on a single NVIDIA V100 (32GB) GPU. FreqMoE consistently maintains two active experts (Topk=2) across all modes.}
    \label{fig:modes_scale}
\end{figure}

\textbf{Scale up frequency Modes sparsely vs densely.}  Figure \ref{fig:modes_scale} demonstrates the trade-offs between model performance and computational cost when scaling frequency modes. Dense FNO shows initial error reduction from modes 4 to 12, but experiences performance degradation with higher modes due to the inherent sparsity of frequency signals. FreqMoE maintains steady improvement through dynamic expert selection. The inference time of dense FNO grows quadratically with modes due to full spectral convolution, while FreqMoE achieves linear complexity by fixing active experts (Topk=2), where modes only affect gating computation. This computational advantage is particularly beneficial for large-scale PDE solving.

\section{Conclusion}
In this paper, we presented FreqMoE, a dynamic frequency enhancement framework that addresses the critical challenge of high-frequency signal loss in Fourier Neural Operators (FNO) through a sparse mixture-of-experts paradigm. By establishing a "Low-Frequency Pretraining, High-Frequency Fine-tuning" strategy, our method efficiently bridges low- and high-frequency learning while maintaining remarkable parameter efficiency. Key innovations include: (1) a frequency-domain MoE architecture that dynamically activates specialized experts to capture sparse high-frequency components, (2) a LoRA-based weight initialization scheme that recycles pretrained FNO weights into high-frequency experts with minimal parameter overhead, and (3) a sparse upward-cycling training framework that achieves up to 47.32× parameter reduction compared to dense FNO variants. Extensive experiments on both regular and irregular grid PDEs demonstrate FreqMoE's superior performance, achieving 16.6\% accuracy improvement on high-resolution tasks (512×512) and significantly enhanced stability in long-term predictions through adaptive frequency allocation.

Despite these advancements, our framework has several limitations. First, the current gating mechanism relies on predefined frequency band partitioning, which may not optimally adapt to all PDE systems’ spectral characteristics. Second, while our parameter-efficient design reduces computational costs, the sparse expert activation introduces minor latency overhead during dynamic routing decisions. Future work will focus on adaptive frequency partitioning strategies and extending the LPHF paradigm to broader operator learning scenarios. We believe FreqMoE establishes a foundational framework for frequency-aware neural PDE solvers, opening new pathways for efficient high-resolution scientific computing.

\clearpage

\section*{Acknowlegement}
This work was supported by the National Science and Technology Major Project(No.2022ZD0117800), and Young Elite Scientists Sponsorship Program by CAST(No.2023QNRC001). This work was also sponsored by CAAI-Huawei MindSpore Open Fund (CAAIXSJLJJ2023MindSpore12) and developed on openl community. Thanks for the computing infrastructure provided by Beijing Advanced Innovation Center for Big Data and Brain Computing.





\bibliographystyle{named}
\bibliography{ijcai25}

\clearpage

\appendix
\section*{Appendix}

\subsection*{A.1 Implementation Details and Additional Results}
\label{app:implentation_details}




To maintain consistency and eliminate potential confounding factors, we employ identical training configurations in all experiments. Specifically, we utilize a cosine learning rate scheduler with a warm-up phase, followed by a sustained period of low learning rate to facilitate extended training epochs, thereby enhancing the stability of model parameters.

In the context of FreqMoE, both the router and all expert networks are jointly optimized to effectively capture the relative importance among experts. During the inference phase, experts are systematically ranked and selectively activated on the basis of their learned importance weights.

The comprehensive set of training hyperparameters is meticulously documented in Table \ref{tab:training_detail}, providing a transparent and reproducible framework for our experimental setup.

\begin{table}[htbp]
\renewcommand{\arraystretch}{1.1}
\centering
\caption{Detailed hyperparameters during model training. }
\label{tab:training_detail}
\begin{tabular}{c|cc}
\toprule
    \multirow{1}{*}{\textbf{General}} & Batch size & 32 \\

    \hline
    \multirow{3}{*}{\textbf{Optimizer}} & Beta1& 0.9\\
    & Beta2& 0.99\\
    & Learning rate & 1e-3 \\
    \hline 
    \multirow{4}{*}{\textbf{Scheduler}} & Min LR ratio & 5e-2 \\
        & Warmup steps & 50 \\
        & Cosine anneling epoches & 70 \\
        & Steady LR epoches & 30 \\

    \hline
    \multirow{1}{*}{\textbf{FNO specifications}} & Width & 32 \\
    
    \hline
    \multirow{4}{*}{\textbf{FreqMoE specifications}} & Modes & 4 \\
    & Width & 32 \\
    & LoRA Rank & 4 \\
    & Max num of experts & 63 \\
    & Sparsity weight & 0.01 \\
\bottomrule
\end{tabular}
\end{table}

\subsection*{A.2 Detailed Problem Description}
\label{app:problem_description}

\subsubsection*{A.2.1 Compressible Navier-Stokes equation}
\label{app:cfd}
The compressible Navier-Stokes equations describe the dynamics of compressible fluid flow through three coupled equations representing mass, momentum, and energy conservation.

Mass conservation:
\begin{equation}
    \partial_t\rho+\nabla\cdot(\rho\mathbf{v}) =0
\end{equation}

Momentum conservation:
\begin{equation}
    \rho(\partial_t\mathbf{v}+\mathbf{v}\cdot\nabla\mathbf{v}) =-\nabla p+\eta\Delta\mathbf{v}+(\zeta+\eta/3)\nabla(\nabla\cdot\mathbf{v})
\end{equation}

Energy conservation:
\begin{equation}
    \partial_t\left[\epsilon+\frac{\rho v^2}{2}\right] +\nabla\cdot\left[\left(\epsilon+p+\frac{\rho v^2}{2}\right)\mathbf{v}-\mathbf{v}\cdot\sigma^{\prime}\right]=0
\end{equation}

In these equations, $\rho$ represents mass density, $\mathbf{v}$ denotes velocity field, $p$ is gas pressure, and $\epsilon=p/(\Gamma-1)$ represents internal energy with $\Gamma = 5/3$. The system also includes viscous effects through the viscous stress tensor $\sigma'$, with $\eta$ and $\zeta$ representing shear and bulk viscosity respectively.

The dataset incorporates two primary types of initial conditions for 2D simulations. The first is a random field initial condition, where density and pressure fields are constructed by superimposing perturbations onto a uniform background. The second type is a turbulent initial condition, characterized by uniform mass density and pressure, where the initial velocity field is described by 
\begin{equation}
    \mathbf{v}(x,t=0)=\sum_{i=1}^n\mathbf{A}_i\sin(k_ix+\phi_i)
\end{equation}
where $n = 4$, and the amplitude $\mathbf{A}_i$ scales with wavenumber as $\bar{v}/|k|^d$, where $d = 2$ for the 2D case. The mean velocity $\bar{v}$ is determined by the product of sound velocity cs and Mach number $M$. To minimize compressibility effects, the compressible component of the velocity field is removed through Helmholtz-decomposition in Fourier space.

The numerical framework employs outgoing boundary conditions, where boundary values are obtained by copying from neighboring cells. This approach effectively allows waves and fluid to escape from the computational domain, making it particularly suitable for astrophysical fluid dynamics simulations. The numerical solution combines a 2nd-order HLLC scheme with MUSCL method for inviscid terms and a central difference scheme for viscous terms. The dataset provides simulations at both standard (128×128) and high (512×512) resolutions, enabling comprehensive evaluation of model performance across different scales and assessment of frequency processing capabilities.

This comprehensive problem setup provides a robust testbed for evaluating numerical methods and machine learning models in their ability to capture complex fluid dynamic phenomena, particularly the rich high-frequency details generated by vortex structures under various flow conditions.

\subsubsection*{A.2.2 Airfoil Problem with Euler's Equation}
\label{app:airfoil}

The Airfoil problem investigates transonic flow over an airfoil governed by the Euler equation system. The system consists of three conservation equations. 

Mass conservation:
\begin{equation}
    \frac{\partial\rho^f}{\partial t}+\nabla\cdot(\rho^f\boldsymbol{v})=0
\end{equation}

Momentum conservation: 
\begin{equation}
    \frac{\partial\rho^f\boldsymbol{v}}{\partial t}+\nabla\cdot(\rho^f\boldsymbol{v}\otimes\boldsymbol{v}+p\mathbb{I})=0
\end{equation}

Energy conservation:
\begin{equation}
    \frac{\partial E}{\partial t}+\nabla\cdot\left((E+p)\boldsymbol{v}\right)=0
\end{equation}

Here, $\rho^f$ represents the fluid density, $v$ denotes the velocity vector, $p$ is the pressure, and $E$ represents the total energy. This formulation neglects viscous effects and focuses on the primary flow dynamics.

The problem is characterized by specific boundary conditions: at the far-field, conditions are set to $\rho_\infty = 1$, $p_\infty = 1.0$, with a Mach number $M_\infty = 0.8$ and angle of attack $AoA = 0$. At the airfoil surface, a no-penetration condition is enforced. The airfoil shape parameterization follows the design element approach, where the initial NACA-0012 profile is mapped onto a cubic design element with 8 control nodes. Shape variations are generated by vertically displacing these control nodes, with displacements drawn from a uniform distribution $U[-0.05, 0.05]$.

The dataset comprises 1000 training samples and 200 test samples, generated using a second-order implicit finite volume solver. The computational domain is discretized using a C-grid mesh with approximately 200 × 50 quadrilateral elements, featuring mesh adaptation near the airfoil but not in shock regions. Each simulation requires approximately 1 CPU-hour to complete. The dataset records both mesh point locations and corresponding Mach number values, which serve as input and output data respectively.

This setup provides a comprehensive test case for evaluating numerical methods and machine learning models in their ability to predict transonic flow fields around varying airfoil geometries, particularly challenging due to the presence of shock waves and complex flow features in the transonic regime.

\subsubsection*{A.2.3 Hyper-elastic Unit Cell Problem}
\label{app:elasticity}

The hyper-elastic problem examines the deformation of a solid body governed by the momentum conservation equation:
\begin{equation}
    \rho^s\frac{\partial^2\boldsymbol{u}}{\partial t^2}+\nabla\cdot\boldsymbol{\sigma}=0
\end{equation}
where $\rho^s$ represents the mass density, $u$ is the displacement vector, and $\sigma$ denotes the stress tensor. The problem considers a unit cell domain $\Omega = [0, 1] \times [0, 1]$ containing a centrally located void with variable radius. The void radius is characterized by $r=0.2+\frac{0.2}{1+\exp(\tilde{r})}$, where $\tilde{r}$ follows a Gaussian process: $\tilde{r}\sim\mathbb{N}(0,4^2(-\nabla+3^2)^{-1})$, constraining the radius between $0.2$ and $0.4$. The unit cell is clamped at its bottom edge with a tension traction $t = [0, 100]$ applied to the top edge. 

The material behavior is described by an incompressible Rivlin-Saunders constitutive model, where the stress-strain relationship is derived from the strain energy density function:
\begin{equation}
    \begin{aligned}
    \boldsymbol{\sigma} & =\frac{\partial w(\boldsymbol{\epsilon})}{\partial\boldsymbol{\epsilon}} \\
    w(\boldsymbol{\epsilon}) & =C_1(I_1-3)+C_2(I_2-3)
    \end{aligned}
\end{equation}
The invariants $I_1$ and $I_2$ are defined as:$I_1 = \mathrm{tr}(C), I_2=\frac{1}{2}[(\mathrm{tr}(C)^2-\mathrm{tr}(C^2)]$, where $C = 2\boldsymbol{\epsilon} + \mathbb{I}$ is the right Cauchy Green stretch tensor. The material parameters are set to $C_1 = 1.863 \times 10^5$ and $C_2 = 9.79 \times 10^3$.

The dataset comprises 1000 training samples and 200 test samples, generated using a finite element solver with approximately 100 quadratic quadrilateral elements. Each simulation requires about 5 CPU seconds to complete. The input data consists of point clouds with approximately 1000 points, and the target output is the stress field. This setup provides a comprehensive test case for evaluating numerical and machine learning methods in predicting stress distributions in hyper-elastic materials with varying geometrical configurations.

\end{document}